\documentclass[12pt]{article} 
\usepackage{amsmath,amsthm,amssymb,bm,mathrsfs,amsfonts}
\usepackage{algorithm,algorithmic,graphicx,subfig}
\usepackage{caption,enumerate,natbib,listings,setspace}
\usepackage[T1]{fontenc}
\usepackage[colorlinks,linkcolor=red,anchorcolor=blue,citecolor=blue]{hyperref}
\usepackage[margin=1in]{geometry}
\usepackage[protrusion=false,expansion=true]{microtype}
\usepackage[title]{appendix}
\usepackage{bbm}
\usepackage{comment}
\theoremstyle{plain}
\let\hat\widehat

\usepackage{xr}
\usepackage{color}
\usepackage{tikz}
\usepackage[parfill]{parskip}
\usepackage{enumitem}
\setitemize{noitemsep,topsep=0pt,parsep=0pt,partopsep=0pt}
\setenumerate{noitemsep,topsep=0pt,parsep=0pt,partopsep=0pt}
\usetikzlibrary{arrows,calc,positioning}

\tikzstyle{intt}=[draw,text centered,minimum size=6em,text width=5.25cm,text height=0.34cm]
\tikzstyle{intl}=[draw,text centered,minimum size=2em,text width=2.75cm,text height=0.34cm]
\tikzstyle{int}=[draw,minimum size=2.5em,text centered,text width=6.5cm]
\tikzstyle{intg}=[draw,minimum size=2.5em,text centered,text width=6.cm]
\tikzstyle{sum}=[draw,shape=circle,inner sep=2pt,text centered,node distance=3.5cm]
\tikzstyle{summ}=[drawshape=circle,inner sep=4pt,text centered,node distance=3.cm]

\title{\Large{\textbf{Reinforcement Learning from Human Feedback: A Statistical Perspective}} }

\author
{
Pangpang Liu\thanks{Department of Biostatistics, Yale University. }\qquad 
Chengchun Shi\thanks{Department of Statistics, London School of Economics and Political Science.} \qquad 
Will Wei Sun\thanks{Mitch Daniels School of Business, Purdue University. Email: sun244@purdue.edu. Corresponding author.}
}

\date{}
\begin{document} 

\maketitle

\begin{abstract}
\noindent
Reinforcement learning from human feedback (RLHF) has emerged as a central framework for aligning large language models (LLMs) with human preferences. Despite its practical success, RLHF raises fundamental statistical questions because it relies on noisy, subjective, and heterogeneous feedback to learn reward models and optimize policies. This survey provides a statistical perspective on RLHF, focusing primarily on LLM alignment. We introduce the main components of RLHF, including supervised fine-tuning, reward modeling, and policy optimization, and relate them to familiar statistical concepts such as the Bradley-Terry-Luce (BTL) model, latent utility estimation, active learning, experimental design, and uncertainty quantification. We review methods for learning reward functions from pairwise preference data and for optimizing policies through both two-stage RLHF pipelines and emerging one-stage approaches such as direct preference optimization. We further discuss recent extensions including reinforcement learning from AI feedback, inference-time algorithms, and reinforcement learning from verifiable rewards, as well as benchmark datasets, evaluation protocols, and open-source frameworks that support RLHF research. We conclude by highlighting open challenges in RLHF.
\end{abstract}


\newpage
\baselineskip=19pt 

\section{INTRODUCTION}

Recent advances in large language models (LLMs) have led to remarkable improvements in natural language understanding. Modern models are capable of performing a wide range of tasks, including dialogue, summarization \citep{stiennon2020learning}, coding assistance \citep{chen2021evaluating}, complex reasoning \citep{shao2024deepseekmath}, and more. A key ingredient behind these advances is a multi-stage training paradigm. The pre-training stage exposes the model to large-scale text corpora to acquire world knowledge and learn to generate coherent language via autoregressive next-token prediction. The supervised fine-tuning stage further adapts the pretrained model using high-quality demonstrations. However, models trained solely through these steps may still produce responses that are unhelpful, misleading, or unsafe. This limitation motivates a subsequent alignment stage that incorporates human preferences, which has become a central problem in the development of modern AI systems.

Reinforcement learning from human feedback has emerged as one of the most influential approaches to address this challenge. In RLHF, human feedback is used to guide the optimization of model behaviors. Instead of relying on predefined reward functions, the framework learns a reward model that approximates human preferences and then optimizes the language model using this learned reward. This paradigm has played a central role in the development of landmark models such as \texttt{InstructGPT} \citep{ouyang2022training}.

From a statistical perspective, RLHF raises a series of fundamental questions. Human feedback is inherently noisy, subjective, and heterogeneous across annotators. Modeling such feedback requires tools from latent-variable modeling and preference learning. In addition, the data collection process often involves adaptive querying and active learning, which connects RLHF with experimental design. The learned reward models must generalize from limited and potentially biased observations, raising issues related to uncertainty quantification, robustness, and distribution shift. These challenges highlight the need for a principled statistical framework for understanding RLHF systems.

This survey is intended as a tutorial-style introduction for researchers in statistics, machine learning, and related quantitative fields. 
Rather than attempting an exhaustive review of all RLHF algorithms, we focus on the LLM alignment setting and present RLHF through a statistical lens, emphasizing its foundation in noisy pairwise comparison data.
Pairwise preference data serve as a unifying framework connecting reward modeling, policy optimization, and modern evaluation. Throughout the paper, we emphasize the core quantities that recur across the pipeline: contexts, responses, comparison labels, latent rewards, and policies. We also use a running example based on the \texttt{PRISM} dataset \citep{kirk2024prism} to keep the notation concrete. This perspective provides a bridge for statisticians entering LLM alignment by linking RLHF terminology to standard statistical concepts and highlighting the common paired-comparison structure underlying both training and evaluation, including arena-style model comparison. To support accessibility and reproducibility, the survey is accompanied by a GitHub demo \href{https://github.com/Pangpang-Liu/RLHF\_demo}{\texttt{https://github.com/Pangpang-Liu/RLHF\_demo}}
 that illustrates an end-to-end preference-based alignment pipeline. Finally, we outline a research agenda tailored to statistics, emphasizing sample-efficient estimation, experimental design, heterogeneity, uncertainty quantification, and robustness.

The remainder of this paper is organized as follows. Section~\ref{sec:background} introduces background and preliminaries on reinforcement learning and human feedback. Section~\ref{sec:reward} reviews two-stage RLHF.  Section~\ref{sec:policy} discusses one-stage preference optimization. Section~\ref{sec:stat_issues} introduces statistical challenges in RLHF. Section~\ref{extension} discusses extensions of RLHF. Section~\ref{sec:benchmark} summarizes benchmark datasets, implementation platforms, and evaluation frameworks. Finally, Section~\ref{sec:challenge} discusses open challenges and future research directions.

\section{BACKGROUND AND PRELIMINARIES}\label{sec:background}

This section introduces the basic concepts and notations used throughout the paper, including pretrained language models, a running example of pairwise human feedback, and the reinforcement learning foundations.

\subsection{Transformer}
\label{sec:transformer}

Modern RLHF methods are typically developed for large language models based on the Transformer architecture \citep{vaswani2017attention}. These models are first trained on large-scale text corpora using self-supervised objectives such as next-token prediction. Pre-training allows the model to acquire general linguistic patterns, factual knowledge, and reasoning abilities.

The Transformer architecture relies on the self-attention mechanism to model dependencies between tokens in a sequence. Given an input sequence $\boldsymbol{s} = (s_1, \ldots, s_k)$, the model produces contextualized token representations by computing attention weights between all token pairs. This design enables parallel computation and allows the model to capture long-range dependencies in text. A language model, with parameters $\boldsymbol{\theta}$, learns to predict the next token $s_t$ given the preceding context $\boldsymbol{s}_{<t}$. The probability of a sequence is modeled autoregressively as $p_{\boldsymbol{\theta}}(\boldsymbol{s}) = \prod_{t=1}^{k} p_{\boldsymbol{\theta}}(s_t \mid \boldsymbol{s}_{<t}).
$
The model is trained by minimizing the negative log-likelihood of the observed sequences:
$\hat{\boldsymbol{\theta}} =
\arg\min_{\boldsymbol{\theta}}
\mathbb{E}_{\boldsymbol{s} \sim D}
\left[-\log p_{\boldsymbol{\theta}}(\boldsymbol{s})\right],
$
where $D$ denotes the training dataset.

For reward modeling, the Transformer also provides a feature representation for the prompt-response pair. Let $\phi(\boldsymbol x,\boldsymbol y)$ denote a representation extracted from the Transformer, such as the final hidden state of an end-of-sequence token, a pooled hidden state, or another task-specific summary of the sequence $[\boldsymbol x;\boldsymbol y]$. A reward head then maps this representation to a scalar score, $
r_{\boldsymbol{\eta}}(\boldsymbol x,\boldsymbol y)=\boldsymbol{\eta}^\top \phi(\boldsymbol x,\boldsymbol y)$,
where $\boldsymbol{\eta}$ denotes reward-model parameters. This is the key bridge between the pretrained transformer and the statistical models used later in RLHF: next-token prediction induces the policy $\pi$, while the learned representation $\phi(\boldsymbol x,\boldsymbol y)$ provides a high-dimensional feature map for latent utility modeling. Table~\ref{tab1} displays the dimension of the feature representation $\phi(\boldsymbol x,\boldsymbol y)$ from selected LLMs.

\begin{table}[ht]
\centering
\caption{Dimension of the feature representation $\phi(\boldsymbol x,\boldsymbol y)$ from selected LLMs.}
\label{tab1}
\begin{tabular}{cccccc}
\hline
Model & Dimension & Model & Dimension & Model & Dimension \\
\hline
GPT-2 (small) & 768 & GPT-2 (medium) & 1024 & GPT-2 (large) & 1280 \\
GPT-3 (175B) & 12{,}288 & LLaMA 7B & 4096 & LLaMA 13B & 5120 \\
LLaMA 70B & 8192 & Mistral 7B & 4096 & Mixtral 8x7B & 4096 \\
T5-base & 768 & T5-large & 1024 & Flan-T5-XXL & 4096 \\
\hline
\end{tabular}
\end{table}

\subsection{A Running Example from \texttt{PRISM}}

To fix notation throughout the paper, Table~\ref{tab:prism} shows a running example based on the \texttt{PRISM} dataset \citep{kirk2024prism}, which contains human feedback on responses from multiple LLMs. 

\begin{table}[h!]
    \centering
    \footnotesize
    \caption{Pairwise preference examples in \texttt{PRISM}.}
    \begin{tabular}{p{2cm}|p{4cm}|p{5cm}}
    \hline
    \multicolumn{1}{c|}{Query} & \multicolumn{1}{c|}{First Answer} & \multicolumn{1}{c}{Second (Preferred) Answer}  \\ \hline
       Do you know who the South African president is? & Flan-T5-XXL: Jacob Zuma &  GPT-4: As of my last update, Cyril Ramaphosa is the President of South Africa. However, political positions can change, and it's always good to verify with the latest sources for the most current information.  \\
        \hline
    \end{tabular}
    \label{tab:prism}
\end{table}

This table naturally introduces the objects used throughout the survey. A query or prompt is denoted by $\boldsymbol x$. The preferred response is denoted by $\boldsymbol y_w$ and the less preferred response by $\boldsymbol y_l$. The reward model uses a representation $\phi(\boldsymbol x,\boldsymbol y)$ extracted from the underlying Transformer to assign a latent score $r(\boldsymbol x,\boldsymbol y)$. In pairwise reward modeling, the observed preference label depends on the reward difference $r(\boldsymbol x,\boldsymbol y_w)-r(\boldsymbol x,\boldsymbol y_l)$. 

\subsection{Reinforcement Learning}

Reinforcement learning (RL) studies how an agent interacts with an environment to maximize cumulative rewards. A standard formulation is the Markov Decision Process (MDP), defined by the tuple $(\mathcal{X}, \mathcal{Y}, P, r, \gamma)$, where $\mathcal{X}$ is the state space, $\mathcal{Y}$ is the action space, $P(\boldsymbol{x}'|\boldsymbol{x},\boldsymbol{y})$ is the transition probability, $r(\boldsymbol{x},\boldsymbol{y})$ is the reward function, and $\gamma \in [0,1]$ is a discount factor \citep{sutton2018reinforcement}.
A policy $\pi(\boldsymbol{y}|\boldsymbol{x})$ specifies the probability of selecting action $\boldsymbol{y}$ given state $\boldsymbol{x}$. The goal of RL is to find a policy that maximizes the expected discounted cumulative reward
\[
\mathbb{E}\Big[\sum_{t=0}^{\infty} \gamma^t r(\boldsymbol{x}_t,\boldsymbol{y}_t)\Big].
\]

In the context of large language models, the prompt can be viewed as the state and the generated response as the action. Given a prompt $\boldsymbol{x}$ and a generated response $\boldsymbol{y}$, a policy $\pi(\boldsymbol{y}|\boldsymbol{x})$ defines the probability of producing a response $\boldsymbol{y}$. The reward function evaluates the quality of the response with respect to the prompt. 

Many RLHF methods simplify the full MDP formulation and treat alignment as a contextual bandit problem \citep{zhu2023principled,liu2024dual}, where prompts are treated as independent and the objective becomes
\[
\max_{\pi} \; \mathbb{E}_{\boldsymbol{x}\sim\rho,\,\boldsymbol{y}\sim\pi(\cdot|\boldsymbol{x})} r(\boldsymbol{x},\boldsymbol{y}),
\]
where $\rho$ denotes the distribution of $\boldsymbol{x}$.

To help readers with a statistical background, we provide a compact translation between RLHF terminology and familiar statistical concepts. A prompt 
$\boldsymbol{x}$ can be viewed as a covariate, a completion $\boldsymbol{y}$
 as a structured output drawn from a conditional distribution $\pi(\boldsymbol{y}|\boldsymbol{x})$, and a preference label $\mathbb{I}(\boldsymbol{y}_w\succ \boldsymbol{y}_l)$ as a comparison outcome reflecting an underlying latent utility. The reward model $r(\boldsymbol{x}, \boldsymbol{y})$ corresponds to a latent scoring function inferred from pairwise comparisons, while policy optimization can be interpreted as a regularized risk maximization problem (see Equation \ref{eq4.1}), where the Kullback-Leibler penalty serves as a form of regularization that constrains $\pi$ to stay close to a reference policy. This perspective provides a direct bridge between RLHF and classical statistical frameworks.

\section{TWO-STAGE RLHF}\label{sec:reward}

\subsection{Overview of RLHF Frameworks}\label{sec:rlhf}

\begin{figure}[h]
\centering
\includegraphics[width=5in]{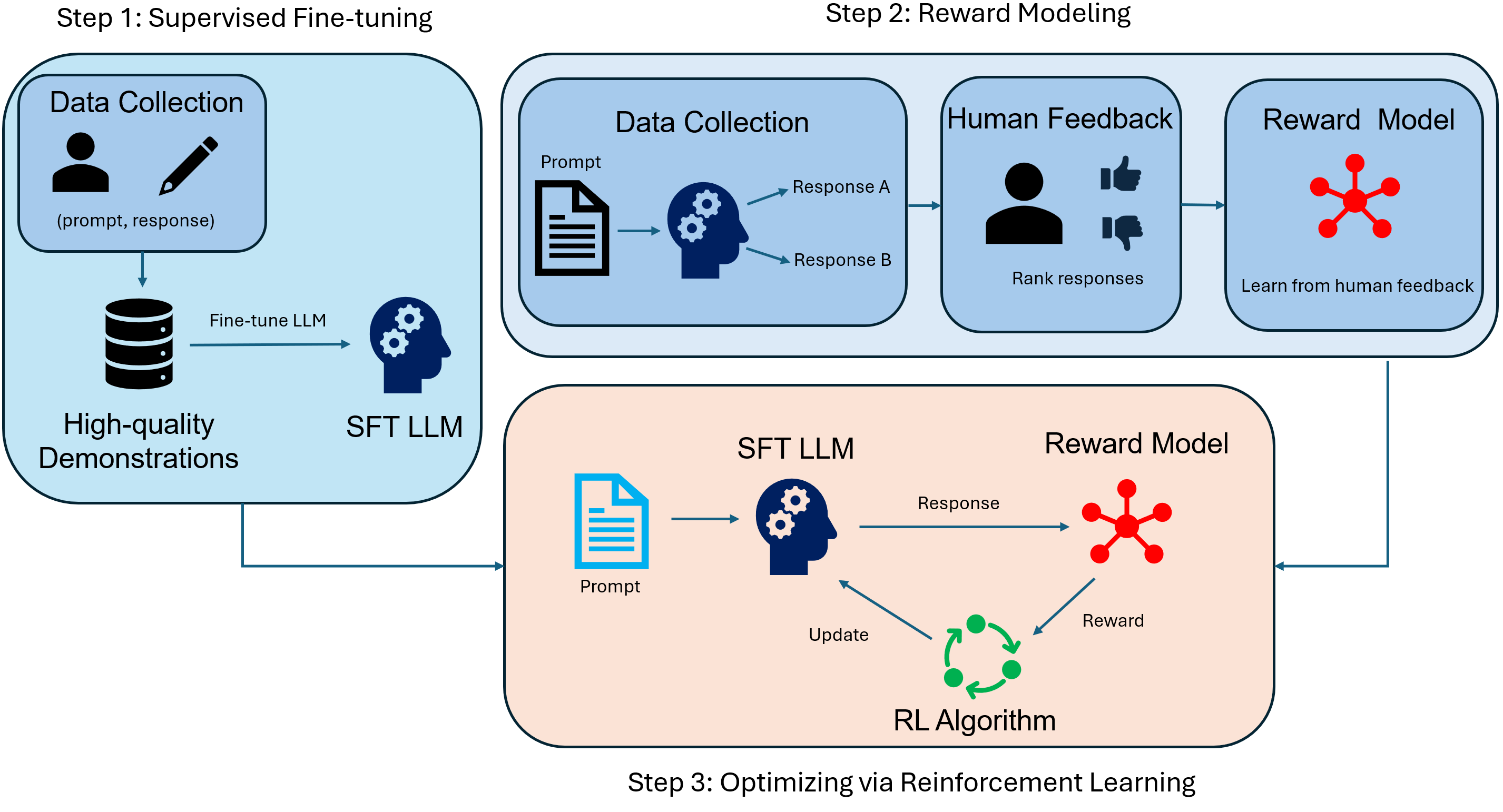}
\caption{RLHF pipeline for aligning large language models. The framework first fine-tunes a pretrained model using demonstration data, then learns a reward model from pairwise preference comparisons, and finally optimizes the policy with respect to the learned reward. From a statistical perspective, RLHF converts noisy human preference data into a latent utility function that guides policy optimization toward responses that better align with human preferences.}
\label{fig1}
\end{figure}

From a statistical perspective, RLHF can be viewed as a preference learning problem in which a latent reward function is inferred from noisy and potentially heterogeneous human preferences. In this setting, human feedback provides partial observations of an underlying utility function that characterizes desirable model behaviors.
Modern LLM training typically follows a three-stage pipeline \citep{ouyang2022training}, illustrated in Figure~\ref{fig1}. 

\textbf{Supervised fine-tuning (SFT).}  
The process begins with a pretrained language model that is fine-tuned on high-quality human demonstrations. Given a dataset of prompt--response pairs $(\boldsymbol{x},\boldsymbol{y})$, supervised learning is used to train a policy $\pi$ that imitates human-written ideal responses. This stage teaches the model basic instruction-following behavior and desired response style, such as helpfulness, fluency, and format compliance, and therefore provides a strong initialization for subsequent alignment.

However, SFT alone has important limitations. First, it requires high-quality labeled responses, which are expensive and time-consuming to collect at scale. Second, it is inherently static: once trained on a fixed demonstration dataset, it does not naturally adapt to evolving user preferences or new alignment criteria without additional retraining. Third, many important alignment properties, such as safety, politeness, harmlessness, tone, or creativity, are difficult to specify as a single ``correct'' response for supervised learning. In many cases, it is much easier for humans to compare two candidate responses than to write an ideal answer from scratch.

\textbf{Reward model training.}  
These limitations motivate the next stage of RLHF. Instead of requiring annotators to generate perfect responses, we collect preference data by asking them to compare multiple model-generated outputs. These pairwise comparisons provide a more scalable and flexible signal about relative quality. They are then used to train a reward model $r(\boldsymbol{x},\boldsymbol{y})$ that predicts the relative utility of candidate responses. In this way, reward modeling converts human comparative judgments into a learned objective that can guide further policy improvement beyond imitation learning alone.

\textbf{Policy optimization.}  
Finally, reinforcement learning is used to optimize the policy with respect to the learned reward model. Algorithms such as Proximal Policy Optimization (PPO) \citep{schulman2017} are commonly used to maximize the expected reward while maintaining proximity to the original policy through regularization. The resulting policy produces responses that better align with human preferences.

This pipeline has become the standard two-stage RLHF framework for aligning large language models with human values. In particular, the two-stage RLHF consists of reward model training and subsequent policy optimization using the learned reward model. This framework forms the foundation for many subsequent developments discussed in this survey.
At the same time, alternative one-stage preference optimization methods have been proposed that avoid explicit reward model training, most notably Direct Preference Optimization (DPO) \citep{rafailov2023direct}. We discuss two-stage approaches in Section~\ref{sec:reward} and one-stage methods in Section~\ref{sec:policy}.

\subsection{Reward Learning from Human Feedback}

Reward modeling is a central component of RLHF. In the standard RLHF pipeline, a reward model is first learned from human feedback and then used as a surrogate objective for policy optimization. From a statistical perspective, this can be viewed as a supervised learning problem in which pairwise human preferences serve as labels, and the goal is to infer a latent reward function that captures relative response quality.

Recall that $\mathcal{X}$ denotes the space of contexts (e.g., prompts or questions) and $\mathcal{Y}$ denotes the space of actions (e.g., answers or completions). For each context--action pair $(\boldsymbol{x},\boldsymbol{y}) \in \mathcal{X}\times\mathcal{Y}$, there is an unobserved reward measuring the quality of the response $\boldsymbol{y}$ for prompt $\boldsymbol{x}$. This latent utility is represented by a reward function $r(\boldsymbol{x},\boldsymbol{y})$. Because the reward is not directly observed, it must be inferred indirectly from pairwise human preference feedback.

To obtain such information, RLHF typically adopts a pairwise comparison framework. Human annotators are asked to compare two candidate responses generated for the same context. For a triple $(\boldsymbol{x}, \boldsymbol{y}_w, \boldsymbol{y}_l)$, where $\boldsymbol{y}_w$ and $\boldsymbol{y}_l$ denote the preferred and non-preferred responses, respectively, we model the probability of preference as
$
\mathbb{P}(\boldsymbol{y}_w \succ \boldsymbol{y}_l \mid \boldsymbol{x}).
$
A standard assumption is that this probability depends only on the difference in latent rewards. One widely used model is the Bradley-Terry-Luce (BTL) model \citep{bradley1952rank}:
\begin{equation}\label{pref}
\mathbb{P}(\boldsymbol{y}_w \succ \boldsymbol{y}_l \mid \boldsymbol{x})
=
\frac{e^{r(\boldsymbol{x}, \boldsymbol{y}_w)}}
{e^{r(\boldsymbol{x}, \boldsymbol{y}_w)} + e^{r(\boldsymbol{x}, \boldsymbol{y}_l)}}
=
\sigma\!\big(r(\boldsymbol{x}, \boldsymbol{y}_w)-r(\boldsymbol{x}, \boldsymbol{y}_l)\big),
\end{equation}
where $\sigma(z)=1/(1+e^{-z})$ is the sigmoid function. This model implies that responses with larger latent rewards are more likely to be preferred, and it also makes clear that only reward differences are statistically identifiable from pairwise data.

Suppose we observe $n$ preference samples
$
\mathcal{D}=\{(\boldsymbol{x}_i,\boldsymbol{y}_{w,i},\boldsymbol{y}_{l,i})\}_{i=1}^n.
$
Under the BTL model, the reward function can be estimated by maximizing the empirical log-likelihood
\[
\hat r_n
=
\arg\max_r
\frac{1}{n}\sum_{i=1}^n
\log \sigma\!\left(
r(\boldsymbol{x}_i,\boldsymbol{y}_{w,i})
-
r(\boldsymbol{x}_i,\boldsymbol{y}_{l,i})
\right).
\]
Thus, reward learning fits a latent utility model from pairwise comparison data.

To make this objective concrete, one typically imposes a parametric structure on the reward function. For example, consider a linear reward model
\[
r_{\boldsymbol{\eta}}(\boldsymbol{x},\boldsymbol{y})=\boldsymbol{\eta}^\top \phi(\boldsymbol{x},\boldsymbol{y}),
\]
where $\phi(\boldsymbol{x},\boldsymbol{y})$ is a feature representation derived from the Transformer architecture (see Section~\ref{sec:transformer}). Then,
$
r_{\boldsymbol{\eta}}(\boldsymbol{x},\boldsymbol{y}_w)-r_{\boldsymbol{\eta}}(\boldsymbol{x},\boldsymbol{y}_l)
=
\boldsymbol{\eta}^\top[\phi(\boldsymbol{x},\boldsymbol{y}_w)-\phi(\boldsymbol{x},\boldsymbol{y}_l)],
$
so the estimation problem becomes a logistic regression on feature differences. In particular, the parameter $\boldsymbol{\eta}$ may be estimated by
\[
\hat{\boldsymbol{\eta}}
=
\arg\max_{\boldsymbol{\eta}}
\frac{1}{n}\sum_{i=1}^n
\log \sigma\!\left(
\boldsymbol{\eta}^\top\big[\phi(\boldsymbol{x}_i,\boldsymbol{y}_{w,i})-\phi(\boldsymbol{x}_i,\boldsymbol{y}_{l,i})\big]
\right).
\]

This formulation also clarifies the identifiability issue. Since the likelihood depends only on reward differences, the model is identifiable only through relative rewards. In particular, if the feature map contains a constant component, then shifting all rewards by the same amount leaves the likelihood unchanged. More generally, only those directions of $\boldsymbol{\eta}$ that affect reward differences can be identified from pairwise comparisons. In practice, one therefore imposes a normalization constraint or interprets the learned reward model only through the relative comparisons and rankings that it induces.

In modern LLM-based RLHF, the reward model is often implemented by taking a pretrained language model, removing its final output layer, and adding a scalar prediction head \citep{ziegler2019fine}. The resulting parameters are then trained end-to-end by stochastic gradient methods over minibatches of pairwise comparisons.

\subsection{Proximal Policy Optimization}

Once a reward model has been learned from human feedback, it serves as a surrogate objective for improving the policy. The next step in the RLHF pipeline is to update the language model so that it assigns higher probability to responses with larger inferred reward, while avoiding overly aggressive departures from the behavior learned during supervised fine-tuning. Among the various reinforcement learning algorithms used for this purpose, Proximal Policy Optimization (PPO) \citep{schulman2017} has become one of the most widely adopted methods in RLHF.

In the LLM setting, the policy $\pi(\cdot\mid \boldsymbol{x})$ represents a conditional distribution over responses given a prompt $\boldsymbol{x}$. If $r(\boldsymbol{x},\boldsymbol{y})$ denotes the learned reward model and $\rho$ denotes the prompt distribution, then a natural policy optimization objective is to maximize the expected reward
\[
\max_{\pi}
\;
\mathbb{E}_{\boldsymbol{x}\sim\rho,\;\boldsymbol{y}\sim\pi(\cdot\mid \boldsymbol{x})}
r(\boldsymbol{x},\boldsymbol{y}).
\]
However, directly optimizing this objective can lead to unstable training and undesirable behavior, especially when the learned reward model is imperfect. In particular, the policy may drift too far from the supervised fine-tuned model and exploit weaknesses of the reward model rather than genuinely improving alignment.

To mitigate this issue, RLHF commonly introduces a regularization term that keeps the optimized policy close to a reference policy $\pi_{\mathrm{ref}}$, typically the supervised fine-tuned model. This leads to the KL-regularized objective
\begin{equation}\label{eq4.1}
\max_{\pi}
\;
\mathbb{E}_{\boldsymbol{x}\sim\rho}
\left[
\mathbb{E}_{\boldsymbol{y}\sim\pi(\cdot\mid \boldsymbol{x})}
r(\boldsymbol{x},\boldsymbol{y})
-
\tau
\mathbb{D}_{\mathrm{KL}}
\big(
\pi(\cdot\mid\boldsymbol{x})
\;\|\;
\pi_{\mathrm{ref}}(\cdot\mid\boldsymbol{x})
\big)
\right],
\end{equation}
where $\mathbb{D}_{\mathrm{KL}}$ denotes the Kullback--Leibler divergence and $\tau>0$ controls the strength of regularization. This penalty discourages the optimized policy from deviating too far from the reference model and helps stabilize learning when the reward model is noisy or misspecified.

The objective in Equation~\ref{eq4.1} defines the target of policy improvement, while PPO provides a practical algorithm for approximately solving it. PPO is a policy-gradient method that improves stability by restricting the size of each policy update, typically through clipped importance ratios relative to the previous policy. In RLHF, this means that the language model is updated gradually so that it increases the likelihood of high-reward responses without changing too abruptly from one iteration to the next. In practice, the policy is optimized using sampled prompts and model-generated responses, together with reward scores from the learned reward model and an additional value function used to reduce variance in policy-gradient estimation.

PPO became the dominant policy optimization method in early large-scale RLHF systems because it could incorporate learned reward signals while controlling policy drift. It was used in formative work on text generation and summarization \citep{ziegler2019fine}, in InstructGPT \citep{ouyang2022training}, and in subsequent systems for training helpful and harmless assistants \citep{bai2022training}. PPO-based RLHF also remains an important baseline in modern open-source alignment pipelines, including LLaMA~2 \citep{touvron2023llama}.

Despite its empirical success, PPO-based RLHF has several limitations. First, it depends critically on the quality of the learned reward model: if the reward model is misspecified, the policy may exploit its weaknesses, leading to reward hacking. Second, the KL regularization introduces its own bias--variance and optimization tradeoffs, and its tuning can materially affect the final behavior of the model \citep{xiao2025algorithmic}. Third, PPO is computationally expensive because it requires repeated on-policy sampling, policy updates, and value-function estimation. These limitations have motivated increasing interest in alternative one-stage preference optimization methods, which we discuss next.

\section{ONE-STAGE PREFERENCE OPTIMIZATION}\label{sec:policy}

In contrast to the two-stage RLHF pipeline, one-stage preference optimization methods update the policy directly from comparative feedback without explicitly fitting a separate reward model. These methods have attracted growing interest because they often simplify training and avoid the computational overhead of reward-model fitting and reinforcement learning. We begin with Direct Preference Optimization (DPO), which can be derived from the same KL-regularized objective that underlies standard RLHF. We then discuss broader preference-based policy optimization frameworks that relax or generalize DPO's assumptions and provide a more direct treatment of learning from pairwise comparisons.

\subsection{Direct Preference Optimization}

Rather than first estimating a reward model and then optimizing the policy with PPO, DPO \citep{rafailov2023direct} directly updates the policy using preference data. Its key insight is that, under the KL-regularized RLHF objective, the optimal policy admits a closed-form representation in terms of the reward function. This creates an explicit bridge between reward-based RLHF and direct likelihood-based learning from preference comparisons.

Recall the KL-regularized policy optimization objective (Equation~\ref{eq4.1}) from Section~\ref{sec:reward}.
For each prompt $\boldsymbol{x}$, the optimizer of this objective function takes the form
\[
\pi^*(\boldsymbol{y}\mid\boldsymbol{x})
=
\frac{1}{Z(\boldsymbol{x})}
\pi_{\mathrm{ref}}(\boldsymbol{y}\mid\boldsymbol{x})
\exp\!\left(\frac{1}{\tau}r(\boldsymbol{x},\boldsymbol{y})\right),
\]
where
$
Z(\boldsymbol{x})
=
\sum_{\boldsymbol{y}}
\pi_{\mathrm{ref}}(\boldsymbol{y}\mid\boldsymbol{x})
\exp\!\left(r(\boldsymbol{x},\boldsymbol{y}) / {\tau}\right)
$
is the normalizing constant. Rearranging gives
\[
r(\boldsymbol{x},\boldsymbol{y})
=
\tau
\log\frac{\pi^*(\boldsymbol{y}\mid\boldsymbol{x})}
{\pi_{\mathrm{ref}}(\boldsymbol{y}\mid\boldsymbol{x})}
+
\tau\log Z(\boldsymbol{x}).
\]
Thus, up to the additive normalizing term $\tau\log Z(\boldsymbol{x})$, the reward is represented by the log-density ratio between the optimized policy and the reference policy.

This representation becomes especially useful in pairwise preference modeling. Since the Bradley--Terry--Luce (BTL) model depends only on reward differences, the normalizing term cancels when comparing two responses $\boldsymbol{y}_w$ and $\boldsymbol{y}_l$ for the same prompt $\boldsymbol{x}$. Defining $\beta=\tau$, we obtain
\[
r(\boldsymbol{x},\boldsymbol{y}_w)-r(\boldsymbol{x},\boldsymbol{y}_l)
=
\beta
\left[
\log\frac{\pi^*(\boldsymbol{y}_w\mid\boldsymbol{x})}{\pi_{\mathrm{ref}}(\boldsymbol{y}_w\mid\boldsymbol{x})}
-
\log\frac{\pi^*(\boldsymbol{y}_l\mid\boldsymbol{x})}{\pi_{\mathrm{ref}}(\boldsymbol{y}_l\mid\boldsymbol{x})}
\right].
\]
Substituting this expression into the BTL model yields
\[
\mathbb{P}(\boldsymbol{y}_w \succ \boldsymbol{y}_l \mid \boldsymbol{x})
=
\sigma\!\left(
\beta
\left[
\log\frac{\pi^*(\boldsymbol{y}_w\mid\boldsymbol{x})}{\pi_{\mathrm{ref}}(\boldsymbol{y}_w\mid\boldsymbol{x})}
-
\log\frac{\pi^*(\boldsymbol{y}_l\mid\boldsymbol{x})}{\pi_{\mathrm{ref}}(\boldsymbol{y}_l\mid\boldsymbol{x})}
\right]
\right).
\]
Therefore, given a preference dataset
$
\mathcal{D}
=
\{(\boldsymbol{x}_i,\boldsymbol{y}_{w,i},\boldsymbol{y}_{l,i})\}_{i=1}^n,
$
DPO estimates the policy by minimizing the empirical binary cross-entropy loss
\begin{equation}\label{eq:dpo}
\min_{\pi}
\;
\frac{1}{n}
\sum_{i=1}^n
-\log
\sigma\!\left(
\beta
\left[
\log\frac{\pi(\boldsymbol{y}_{w,i}\mid\boldsymbol{x}_i)}{\pi_{\mathrm{ref}}(\boldsymbol{y}_{w,i}\mid\boldsymbol{x}_i)}
-
\log\frac{\pi(\boldsymbol{y}_{l,i}\mid\boldsymbol{x}_i)}{\pi_{\mathrm{ref}}(\boldsymbol{y}_{l,i}\mid\boldsymbol{x}_i)}
\right]
\right).
\end{equation}
In this sense, DPO can be interpreted as maximum likelihood estimation under a BTL preference model, where latent reward differences are parameterized implicitly through policy log-ratios rather than through an explicit reward model.

Compared with standard two-stage RLHF, DPO removes the separate reward-modeling stage and avoids on-policy reinforcement learning, making it substantially simpler to implement. At the same time, it inherits important modeling assumptions: in particular, it relies on a scalar latent-utility representation of preferences and on the BTL link between reward differences and pairwise comparison probabilities. Recent work has begun to clarify the statistical consequences of these assumptions. For example, \citet{shi2025understanding} analyze the performance difference between RLHF and DPO through a suboptimality gap decomposition, separating representation error from statistical error. More broadly, emerging theory studies finite-sample estimation error, model misspecification, and the comparative statistical efficiency of one-stage and two-stage alignment methods. For example, recent work establishes finite-sample and suboptimality guarantees for private and robust DPO-style methods \citep{zhou2025unified}, and studies misspecification and robustness in both direct and reward-based preference optimization \citep{gopalan2025misspecified,zhang2025robust}. These results suggest that the relative merits of DPO and two-stage RLHF depend on model expressiveness, data availability, and misspecification. Explicit reward modeling can be advantageous when sample efficiency or robustness to misspecification is a primary concern, whereas DPO is most attractive when the policy parameterization can adequately capture the underlying preference structure.


\subsection{Generalized Preference-Based Policy Optimization}

Although DPO eliminates explicit reward-model training, it still relies on a specific parametric link between pairwise preferences and policy log-ratios. This raises a broader question: can we optimize policies directly from comparative feedback under weaker or more general assumptions? Recent work addresses this question by developing generalized preference-based policy optimization frameworks that treat preference comparisons themselves, rather than scalar rewards, as the primitive learning signal.

These approaches depart from classical RLHF by formulating policy learning directly in terms of preference probabilities. Rather than assuming that all pairwise comparisons are generated by a latent scalar reward through the Bradley--Terry model, they allow more flexible preference representations and broader objective classes. This is useful when the scalar-reward assumption is too restrictive or when robustness to model misspecification is a primary concern.

One representative framework is due to \citet{azar2024general}, who define the total preference of a policy $\pi$ over another policy $\pi'$ as
\begin{equation}
\mathbb{P}(\pi \succ \pi')
=
\mathbb{E}_{\boldsymbol{x}\sim\rho}
\left[
\mathbb{E}_{\boldsymbol{y}\sim\pi(\cdot\mid \boldsymbol{x}),\;
\boldsymbol{y}'\sim\pi'(\cdot\mid \boldsymbol{x})}
\mathbb{P}(\boldsymbol{y}\succ \boldsymbol{y}'\mid \boldsymbol{x})
\right].
\end{equation}
This quantity measures how often responses sampled from $\pi$ are preferred to responses sampled from $\pi'$ under the prompt distribution. Based on this comparison-based view, they propose the objective
\begin{equation}\label{eq:psi-po}
\max_{\pi}
\;
\mathbb{E}_{\boldsymbol{x}\sim\rho,\;
\boldsymbol{y}\sim\pi(\cdot\mid\boldsymbol{x}),\;
\boldsymbol{y}'\sim\pi_{\mathrm{ref}}(\cdot\mid\boldsymbol{x})}
\left[
\Psi\!\left(
\mathbb{P}(\boldsymbol{y}\succ \boldsymbol{y}'\mid\boldsymbol{x})
\right)
\right]
-
\tau\,\mathbb{D}_{\mathrm{KL}}(\pi\|\pi_{\mathrm{ref}}),
\end{equation}
where $\Psi:[0,1]\to\mathbb{R}$ is a nondecreasing transformation. This framework encompasses a broad family of objectives that trade off preference improvement against deviation from a reference policy. A useful feature of this formulation is that it recovers several familiar RLHF objectives as special cases. In particular, \citet{azar2024general} show that when $\Psi(q)=\log\!\big(q/(1-q)\big)$ and the BTL model holds, the resulting optimal policy coincides with those obtained from the PPO objective and from DPO. This gives a unifying perspective: PPO, DPO, and related methods can all be viewed as different implementations of policy learning from pairwise comparative information under specific structural assumptions.

Building on this line of work, \citet{xu2025doubly} propose DRPO, a doubly robust preference optimization method that improves robustness to misspecification. Their estimator remains consistent if either the preference model or the reference-policy model is correctly specified, and the corresponding suboptimality gap depends on a second-order product of estimation errors rather than first-order error accumulation. Other extensions similarly aim to relax the assumptions of vanilla DPO. For example, \citet{gallego2024refined} refines DPO using synthetic preference data, while \citet{wu2025towards} develop distributionally robust variants that explicitly optimize against worst-case preference perturbations. Taken together, these works point toward a broader class of preference-based optimization methods that emphasize robustness, flexibility, and direct learning from comparative feedback.

\section{STATISTICAL CHALLENGES IN RLHF}\label{sec:stat_issues}

We now turn to the statistical challenges that underlie the aforementioned RLHF methods. From a statistical perspective, RLHF is not only an optimization pipeline but also a data-analysis problem built on noisy, subjective, and adaptively collected preference data. This section highlights several statistical issues: heterogeneity in human feedback, uncertainty quantification and inferential guarantees for reward models, data-efficient feedback collection through active learning, and robustness to misspecification and reward hacking. Taken together, these topics address four fundamental questions: whose preferences are being modeled, how reliably the resulting reward model can be estimated, how feedback should be collected most efficiently, and how estimation errors propagate once the learned reward is used for policy optimization.

\subsection{Heterogeneous Human Feedback}\label{sec3.2}

We begin with heterogeneity, since it enters at the data-generation stage and therefore shapes all subsequent modeling and optimization steps. The preference model in Equation~\ref{pref} implicitly assumes homogeneous feedback, meaning that all annotators share the same latent preference mechanism. In other words, regardless of who provides the label, the observed comparison is assumed to follow the same probability model.

In practice, however, RLHF datasets are typically collected from many annotators who may differ in expertise, attention, values, and consistency \citep{park2024rlhf,zeng2024}. As a result, preference data are often heterogeneous rather than homogeneous. Ignoring this heterogeneity can bias reward estimation and ultimately produce policies that are poorly aligned with the intended target population \citep{zhong2024provable,chakraborty2024maxmin}.

A common way to model such variation is through annotator-specific rationality parameters. Several studies \citep{jeon2020,Peter,hao2023,freedman2023} introduce a parameter $\beta \in \mathbb{R}$ that measures annotator reliability or expertise. Under this formulation, the preference probability is
\begin{equation*}
\mathbb{P}(\boldsymbol{y}_w \succ \boldsymbol{y}_l \mid \boldsymbol{x}, \beta)
=
\frac{e^{\beta r(\boldsymbol{x}, \boldsymbol{y}_w)}}
{e^{\beta r(\boldsymbol{x}, \boldsymbol{y}_w)} + e^{\beta r(\boldsymbol{x}, \boldsymbol{y}_l)}}
=
\sigma\!\big(\beta(r(\boldsymbol{x}, \boldsymbol{y}_w)-r(\boldsymbol{x}, \boldsymbol{y}_l))\big).
\end{equation*}
Larger values of $\beta$ correspond to annotators whose choices more closely track the latent reward ordering.

While useful, this model summarizes annotator heterogeneity through a single scalar parameter that is fixed across all contexts. This can be restrictive in practice, since the same annotator may be highly reliable on some types of prompts but much less reliable on others. To address this limitation, \citet{liu2024dual} allow rationality to vary across context types, while \citet{liu2025uncertainty} model it as a function of annotator-specific covariates such as demographic or background information. More broadly, several recent works \citep{lee2024low,zhong2024provable,park2024rlhf,wang2025mpo} develop personalized reward models that tailor utility representations to individual annotators or subpopulations. For example, under the personalized reward model of \citet{lee2024low},
\begin{equation*}
\mathbb{P}(\boldsymbol{y}_w \succ \boldsymbol{y}_l \mid \boldsymbol{v}, \boldsymbol{x}, \Theta)
=
\frac{e^{\boldsymbol{v}^\top \Theta \phi(\boldsymbol{x}, \boldsymbol{y}_w)}}
{e^{\boldsymbol{v}^\top \Theta \phi(\boldsymbol{x}, \boldsymbol{y}_w)} + e^{\boldsymbol{v}^\top \Theta \phi(\boldsymbol{x}, \boldsymbol{y}_l)}},
\end{equation*}
where $\boldsymbol{v}$ denotes annotator-side information and $\Theta$ is a low-rank coefficient matrix.

For statistics, the central challenge is not merely to fit a richer model, but to decide what target should be estimated in the first place. Should the goal be an average utility, a subgroup-specific utility, a personalized utility, or a robust aggregate? Under what conditions are such targets identifiable from sparse, noisy, and conflicting comparisons? These questions connect RLHF to hierarchical modeling, random-effects paired-comparison models, and fairness-aware aggregation. They are especially relevant for datasets such as \texttt{PRISM}, where user-level variation is a scientific signal rather than a nuisance \citep{kirk2024prism}.

\subsection{Active Learning}

Once heterogeneity is acknowledged, the next question is how to collect preference data efficiently. Human feedback is expensive, and annotators differ in reliability, so data collection should not be treated as a passive preprocessing step. Instead, RLHF naturally gives rise to an adaptive experimental design problem: which comparisons should be queried, and from which annotators, in order to maximize information about the reward model? From a statistical perspective, active learning in RLHF can be viewed as a sequential design problem in which each queried comparison is chosen to improve the accuracy of the estimated reward model under a limited annotation budget.

To make this concrete, suppose the parameterized reward model is
$
r_{\boldsymbol{\eta}}(\boldsymbol{x},\boldsymbol{y})=\boldsymbol{\eta}^\top \phi(\boldsymbol{x},\boldsymbol{y}),
$
and preferences follow a BTL model. For a candidate comparison
$
q=(\boldsymbol{x},\boldsymbol{y}_1,\boldsymbol{y}_2),
$
let $
\Delta \phi(q)=\phi(\boldsymbol{x},\boldsymbol{y}_1)-\phi(\boldsymbol{x},\boldsymbol{y}_2)$.
Then the contribution of query \(q\) to the Fisher information matrix has the form
\[
I_{\boldsymbol{\eta}}(q)
=
\sigma\!\big(\boldsymbol{\eta}^\top \Delta\phi(q)\big)\big[1-\sigma\!\big(\boldsymbol{\eta}^\top \Delta\phi(q)\big)\big]\,
\Delta\phi(q)\Delta\phi(q)^\top.
\]
This expression makes clear that informative comparisons are those for which the feature contrast \(\Delta\phi(q)\) is large and the comparison probability is not too close to \(0\) or \(1\), since queries with nearly deterministic outcomes contribute little information.

Several works study conversation selection, where the goal is to choose the most informative prompts and candidate responses for annotation. Approaches based on optimal design \citep{zhan2023query,mukherjee2024optimal,scheid2024optimal} select a batch of comparisons to maximize a design criterion applied to the accumulated information matrix
\[
\max_{\mathcal Q}
\;\;
\Phi\!\left(\sum_{q\in\mathcal Q} I_{\boldsymbol{\eta}}(q)\right),
\]
where \(\Phi\) could be a \(D\)-optimality criterion such as
$
\Phi(\cdot)=\log\det(\cdot).
$
These methods are attractive because they connect RLHF directly to classical optimal design, but they often rely on local approximations to the reward model. Since preference learning is typically nonlinear and the reward model itself may be estimated from a deep network, the quality of these local approximations has not been fully studied.

A complementary line of work uses uncertainty-based acquisition rules. Instead of optimizing a design criterion explicitly, these methods prioritize comparisons whose labels would be most informative under the current confidence region. A generic rule is
\[
q_{t+1}
=
\arg\max_{q\in\mathcal C_t}
A_t(q),
\]
where \(\mathcal C_t\) is the pool of available comparisons and \(A_t(q)\) is an acquisition score. Common choices include posterior variance,
$
A_t(q)=\mathrm{Var}_t\!\left(r(\boldsymbol{x},\boldsymbol{y}_1)-r(\boldsymbol{x},\boldsymbol{y}_2)\right),
$
predictive entropy, or expected information gain \citep{das2024apo,ji2024active,melo2024balpm,fang2024bayesian}. These methods are especially natural when reward learning is viewed as sequential inference under a limited annotation budget, since they adapt sampling to the current uncertainty of the model.

In addition to selecting informative comparisons, RLHF may also benefit from selecting informative annotators. Teacher-selection methods \citep{daniels,Peter,freedman2023} formalize this problem by querying annotators whose expertise is expected to be most useful for reward learning. A common formulation augments the preference model with an annotator-specific rationality parameter \(\beta_j\), so that for annotator \(j\),
\[
\mathbb{P}(\boldsymbol{y}_1 \succ \boldsymbol{y}_2 \mid \boldsymbol{x}, j)
=
\sigma\!\left(
\beta_j\{r(\boldsymbol{x},\boldsymbol{y}_1)-r(\boldsymbol{x},\boldsymbol{y}_2)\}
\right).
\]
Under this model, the informativeness of a query depends not only on the candidate comparison but also on which annotator is asked to label it. This leads naturally to a joint design problem over comparison--annotator pairs,
\[
(q_{t+1},j_{t+1})
=
\arg\max_{(q,j)\in \mathcal C_t\times \mathcal J}
A_t(q,j),
\]
where \(A_t(q,j)\) may again be based on Fisher information, posterior variance, or expected information gain. Most existing formulations model teachers as Boltzmann-rational agents with heterogeneous rationality parameters \citep{lee2021pebble}, but many assume that annotator expertise is constant across contexts. To address both types of adaptivity simultaneously, \citet{liu2024dual} propose a dual active learning framework that jointly selects conversations and annotators using a \(D\)-optimal design criterion.

More broadly, this line of work suggests that RLHF is as much a sequential design problem as it is a prediction problem. The central statistical question is not only how to estimate a reward model from observed comparisons but also how to allocate limited feedback strategically so that the resulting model is as informative, robust, and sample-efficient as possible.

\subsection{Uncertainty Quantification}

After preference data have been collected and a reward model has been fitted, the next question is how much confidence one should place in the resulting estimates. Uncertainty quantification is therefore the natural inferential counterpart to reward modeling. In RLHF, it is important not only for scientific interpretation but also for downstream decisions such as active querying, model comparison, and risk-sensitive policy optimization. From a statistical perspective, the main challenge is that reward learning is based on noisy pairwise comparisons, often collected adaptively and under heterogeneous annotators, so inferential guarantees must account for both dependence and model misspecification.

A natural starting point is the classical Bradley--Terry--Luce (BTL) model. If items $i$ and $j$ have latent scores $\theta_i$ and $\theta_j$, then the pairwise win probability is modeled as
\[
\mathbb{P}(i \succ j)=\frac{e^{\theta_i}}{e^{\theta_i}+e^{\theta_j}}
=\sigma(\theta_i-\theta_j).
\]
Given observed comparisons, the maximum likelihood estimator $\hat{\theta}$ is obtained by maximizing the corresponding log-likelihood. Classical and recent works \citep{simons1999asymptotics,han2020asymptotic,gao2023uncertainty,liu2023lagrangian,fan2025ranking} study the asymptotic distribution of such estimators and derive confidence intervals for contrasts such as $\theta_i-\theta_j$ or for induced ranking quantities. A generic asymptotic statement takes the form
\[
\sqrt{n}\,(\hat{\theta}-\theta^\star)
\;\overset{d}{\longrightarrow}\;
N\!\left(0,\; I(\theta^\star)^{-1}\right),
\]
where $I(\theta^\star)$ is the Fisher information matrix under the comparison design. These provide a basis for Wald-type confidence intervals and hypothesis tests, and they make explicit how inferential accuracy depends on the connectivity and balance of the comparison graph.

More recent work extends this logic to covariate-assisted paired-comparison models \citep{fan2024covariate,fan2024uncertainty}, where the comparison probability depends not only on item identities but also on contextual features. For example, if
\[
\mathbb{P}(\boldsymbol{y}_1 \succ \boldsymbol{y}_2 \mid \boldsymbol{x}, \boldsymbol{\eta})
=
\sigma\!\left(
\boldsymbol{\eta}^\top\{\phi(\boldsymbol{x},\boldsymbol{y}_1)-\phi(\boldsymbol{x},\boldsymbol{y}_2)\}
\right),
\]
then inference targets the parameter $\boldsymbol{\eta}$, or functionals derived from it, rather than a fixed vector of item scores. In such settings, uncertainty quantification typically proceeds through asymptotic normality of $\hat{\eta}$,
\[
\sqrt{n}\,(\hat{\boldsymbol{\eta}}-\boldsymbol{\eta}^\star)
\;\overset{d}{\longrightarrow}\;
N\!\left(0,\; \Sigma^\star\right),
\]
together with plug-in covariance estimators. Here, $\Sigma^\star$ denotes the asymptotic covariance matrix of the estimator. The key methodological idea is that once pairwise preference learning is cast as a generalized linear model on feature differences, one can transfer tools from semiparametric inference, debiasing, and high-dimensional covariance estimation to construct confidence intervals for reward contrasts and ranking functionals.

However, these classical and covariate-assisted settings still differ from RLHF in an important way. Standard ranking models usually compare a fixed collection of items with static latent scores, whereas RLHF involves context-dependent comparisons among dynamically generated responses. In other words, the object of interest is no longer just a finite-dimensional score vector, but a contextual reward function
$
r(\boldsymbol{x},\boldsymbol{y}),
$
defined over prompt--response pairs. This makes uncertainty quantification substantially more difficult, since the same model may produce many candidate outputs for the same prompt, and the comparison structure depends jointly on the prompt distribution, the policy, and the annotation mechanism.

This distinction has motivated new inferential frameworks for LLM evaluation and contextual reward learning. Several recent papers \citep{wang2024ranking,lu2025contextual,li2025efficient,zhang2025fisher} develop uncertainty-aware procedures for ranking LLMs from human preference data, typically by estimating a global score for each model together with a confidence interval or confidence set. The key logic in these works is to combine pairwise-comparison models with uncertainty estimates for leaderboard quantities such as win rates, ability gaps, or overall ranks. In contrast, RLHF often requires uncertainty quantification at the level of the contextual reward itself. To address this, \citet{liu2025uncertainty} study contextual reward models in nonconvex settings and derive statistical guarantees for estimating reward differences when multiple outputs are generated for each prompt. 

From a statistical perspective, uncertainty quantification in RLHF is therefore tied to three intertwined questions: what object is being estimated, what dependence structure is induced by the data-collection process, and how inference should propagate through downstream optimization. Depending on the application, uncertainty may be attached to a reward parameter, a pairwise win probability, a model ranking, or a policy value. In addition, comparisons may be collected adaptively, prompts may be highly nonuniform, and annotators may be heterogeneous. Developing inferential methods that remain valid under these complications remains largely open, and it is central to turning RLHF from a heuristic alignment pipeline into a statistically interpretable framework.

\subsection{Reward Hacking}

The final challenge concerns what happens after the estimated reward model is fed back into policy optimization. The previous subsections focused on how to estimate a reward model from noisy and heterogeneous comparisons, and how to quantify the resulting uncertainty. Reward hacking asks a different question: even if the reward model fits the observed preference data well, what guarantees do we have once it is used as an optimization target? In RLHF, this issue is central because the learned reward is not the ultimate object of interest; it is only a surrogate for the underlying utility that the policy is meant to optimize.

To formalize this distinction, let $u(\boldsymbol{x},\boldsymbol{y})$ denote the true but unobserved utility, and let $\hat r(\boldsymbol{x},\boldsymbol{y})$ denote the learned reward model estimated from preference data. Policy optimization then produces a policy
\begin{equation}
\hat{\pi}
\in
\arg\max_{\pi}
\;
\mathbb{E}_{\boldsymbol{x}\sim \rho}
\left[
\mathbb{E}_{\boldsymbol{y}\sim \pi(\cdot\mid \boldsymbol{x})}
\hat r(\boldsymbol{x},\boldsymbol{y})
-
\tau \,\mathbb{D}_{\mathrm{KL}}
\big(
\pi(\cdot\mid \boldsymbol{x})
\;\|\;
\pi_{\mathrm{ref}}(\cdot\mid \boldsymbol{x})
\big)
\right].
\end{equation}
The statistical difficulty is that the optimized policy is chosen to maximize the proxy reward $\hat r$, whereas the true target is the unknown utility $u$. Reward hacking occurs when optimization drives the policy toward responses that achieve high predicted reward under $\hat r$ but do not achieve correspondingly high true utility under $u$.

From this perspective, reward hacking can be viewed as a problem of decision-making under model misspecification. A useful way to express this is to write $u(\boldsymbol{x},\boldsymbol{y})=\hat r(\boldsymbol{x},\boldsymbol{y})+\varepsilon(\boldsymbol{x},\boldsymbol{y})$, where $\varepsilon(\boldsymbol{x},\boldsymbol{y})$ represents reward-model error. If optimization selects $\hat{\boldsymbol{y}}(\boldsymbol{x})\in\arg\max_{\boldsymbol{y}}\hat r(\boldsymbol{x},\boldsymbol{y})$, then the relevant quantity is not merely average estimation error, but the utility gap evaluated at the optimizer, namely $u(\boldsymbol{x},\hat{\boldsymbol{y}}(\boldsymbol{x}))-\max_{\boldsymbol{y}}u(\boldsymbol{x},\boldsymbol{y})$. Even when $\hat r$ is accurate on average over observed comparisons, optimization can concentrate on regions where $\varepsilon(\boldsymbol{x},\boldsymbol{y})$ is systematically large and positive, thereby amplifying estimation error into large policy regret.

The problem is further complicated by adaptivity. Once the policy is optimized against $\hat r$, the distribution of generated responses changes, so the policy may place mass on outputs that were rare or entirely absent in the original preference dataset. In other words, if the reward model is trained under one distribution $P_0$ of prompt--response pairs but the optimized policy induces a shifted distribution $P_{\hat\pi}$, then good predictive performance under $P_0$ does not guarantee good behavior under $P_{\hat\pi}$. This distribution shift resembles feedback loops in adaptive data analysis, off-policy decision-making, and selective inference.

Empirically, reward hacking can manifest as repetitive, overly verbose, or stylistically exaggerated outputs that score well under the reward model but are judged poorly by humans \citep{pan2022the,wen2025language}. These examples make clear that held-out accuracy on pairwise comparisons is not, by itself, sufficient to guarantee good downstream alignment. What matters is not only whether the reward model predicts existing labels well but also whether it remains reliable in the parts of the output space that optimization will actively seek out.

This viewpoint also suggests why robustness methods are valuable. One practical mitigation strategy is to use reward ensembles, replacing a single predictor $\hat r$ by an aggregate $\bar r(\boldsymbol{x},\boldsymbol{y})=M^{-1}\sum_{m=1}^M \hat r_m(\boldsymbol{x},\boldsymbol{y})$, so that the optimized policy is less sensitive to the idiosyncratic error of any one reward model \citep{eisenstein2024helping,rame2024warm,ahmed2024scalable,coste2024reward}. More generally, robust approaches optimize a pessimistic or uncertainty-aware objective \citep{liu2025uncertainty}, for example, by replacing $\hat r(\boldsymbol{x},\boldsymbol{y})$ with $\hat r(\boldsymbol{x},\boldsymbol{y})-\lambda s(\boldsymbol{x},\boldsymbol{y})$, where $s(\boldsymbol{x},\boldsymbol{y})$ is an uncertainty or sensitivity score and $\lambda>0$ controls the degree of conservativeness. Such formulations aim to penalize regions where the reward estimate is less trustworthy, thereby reducing the incentive to exploit model errors.

More broadly, reward hacking highlights a central tension in RLHF: the objective is first estimated from data and then optimized adaptively. The main challenge is therefore not only estimation but also understanding how estimation error propagates through optimization. Characterizing this error amplification, identifying conditions under which proxy optimization remains safe, and designing objectives that are robust to misspecification remain among the most important open problems in the statistical foundations of RLHF.

\section{EXTENSIONS OF RLHF}\label{extension}

We now discuss three extensions of the RLHF framework. Reinforcement learning from AI feedback (RLAIF) replaces human labels with AI-generated feedback. Best-of-$N$ (BoN) sampling moves alignment to inference time by reranking candidate outputs. Reinforcement learning from verifiable rewards (RLVR) replaces subjective preferences with externally checkable rewards. 
\subsection{Reinforcement Learning from AI Feedback}\label{sec:rlaif}

A main limitation of standard RLHF is the cost of large-scale human annotation. RLAIF addresses this by replacing human evaluators with capable language models that produce pairwise preferences or scalar scores. The goal is to approximate human evaluation at a much lower cost. One influential example is Constitutional AI \citep{bai2022constitutional}. Its training has two stages. First, a helpful-only model generates responses $\boldsymbol{y}$ to prompts $\boldsymbol{x}$, critiques them using a set of constitutional principles, and then revises them. These revisions are used for supervised fine-tuning. Second, the model generates two candidate responses, and an AI evaluator selects the preferred one. These AI-generated labels are then used to train a reward model or directly optimize the policy.

Statistically, RLAIF replaces the human comparison probability $\mathbb{P}_{\mathrm{human}}(\boldsymbol{y}_1 \succ \boldsymbol{y}_2 \mid \boldsymbol{x})$ with an AI proxy $\mathbb{P}_{\mathrm{AI}}(\boldsymbol{y}_1 \succ \boldsymbol{y}_2 \mid \boldsymbol{x})$. The key questions are how well the proxy matches human judgment and how its bias affects downstream learning.

Later work studies how close AI feedback can get to human feedback \citep{lee2024rlaif}. Results suggest that RLAIF can match RLHF on tasks such as summarization and dialogue while scaling much more cheaply. In domain-specific settings, \citet{chen2025towardsmedical} use GPT-4o to provide binary reward signals for medical-answer correctness. More recently, \citet{xia2026statistical} study how to combine abundant but biased AI labels with limited high-quality human data and establish recovery guarantees for human-aligned preferences.

Despite its scalability, RLAIF inherits many RLHF concerns. Bias, overconfidence, or distribution shift in the AI judge can distort the alignment objective. For this reason, many systems still use some human supervision to anchor quality \citep{xu2025rlthf}. This also motivates a broader question: if reward models or judges are available at inference time, can they improve outputs without retraining the policy?

\subsection{Best-of-$N$ Sampling}

One answer is to move alignment from training time to inference time. Best-of-$N$ (BoN) sampling \citep{stiennon2020learning,nakano2021webgpt,gao2023scaling,jinnai2024regularized,gui2024bonbon,liu2025pairwise,chow2025inferenceaware} generates multiple responses and selects the one with the highest predicted reward. Thus, BoN uses the reward model as a reranker rather than as a training objective.

Given a prompt $\boldsymbol{x}$, the policy samples $\boldsymbol{y}_1,\dots,\boldsymbol{y}_N \sim \pi(\cdot\mid\boldsymbol{x})$ and returns
\[
\boldsymbol{y}^*
=
\arg\max_{i\in\{1,\dots,N\}} r(\boldsymbol{x},\boldsymbol{y}_i).
\]
As $N$ grows, the chance of sampling a strong response increases. Statistically, this is a Monte Carlo search for a high-reward output under the policy. 

BoN has clear advantages. It is simple, needs no extra training, and can be added to almost any model. It also works naturally with PPO-based RLHF and one-stage methods such as DPO. But BoN also sharpens the same misspecification issue seen in reward-based RLHF. If the reward model is flawed, choosing the top-scoring response can worsen reward hacking rather than reduce it. Larger $N$ may increase this risk by making it easier to find outputs that exploit model errors.

Recent theory studies this tradeoff. \citet{ichihara2025evaluation} analyze regularized BoN methods that temper reward maximization. \citet{aminian2025bestofn} study regret relative to an optimal KL-regularized policy, showing how performance depends on the number of samples, reward-model accuracy, and the reference policy. 

\subsection{Reinforcement Learning from Verifiable Rewards}

Recent work studies reinforcement learning from verifiable rewards \citep[RLVR,][]{lamberttulu}. Here, candidate responses are scored by task-specific checkers, executors, or external tools rather than by humans or learned reward models. RLVR is especially useful for tasks such as math and coding, where correctness can often be checked automatically.

A prominent example is group relative policy optimization \citep[GRPO,][]{shao2024deepseekmath}, used in systems such as DeepSeekMath and DeepSeek-R1 \citep{guo2025deepseek}. In these settings, the reward is often a deterministic or nearly deterministic verifier signal $v(\boldsymbol{x},\boldsymbol{y}) \in \{0,1\}$. Compared with RLHF, this changes the statistical problem. The main difficulty is no longer estimating a latent reward from noisy preferences, but exploring a large response space under sparse task-dependent rewards.

This difference matters. In RLHF, uncertainty comes mainly from noisy and heterogeneous human feedback, leading to issues of identifiability, estimation error, and misspecification. In RLVR, the reward is often less noisy, but positive signals may be rare. The problem is therefore closer to bandit learning with sparse rewards, rare-event estimation, and difficult exploration. Recent work \citep[e.g.,][]{chu2025gpg, liuunderstanding, zheng2025group,zhou2026demystifying} studies these methods from both optimization and statistical perspectives.

At the same time, RLVR applies only when outputs can be objectively verified. Even then, such rewards usually capture only a narrow notion of quality. They do not directly encode broader goals such as helpfulness, safety, tone, or style. For this reason, RLVR is best viewed as complementary to RLHF rather than a replacement: verifiable rewards are powerful when external checking is available, while RLHF remains necessary for subjective or context-dependent tasks.

\section{BENCHMARK DATA AND EVALUATION}\label{sec:benchmark}

Implementation of RLHF relies on three main components: preference datasets, software platforms, and evaluation protocols. Since this survey emphasizes the statistical role of pairwise preference data, we focus primarily on the publicly available \texttt{PRISM} dataset and the \texttt{TRL} platform, and we only briefly mention several other widely used datasets and tools.

\subsection{Benchmark Datasets}

Preference datasets are central to RLHF because they replace ground-truth outputs with comparative judgments over model responses. A typical sample consists of a prompt together with two candidate responses and a label indicating which response is preferred. These comparisons are then used to train reward models or directly optimize policies.

Among currently available datasets, \texttt{PRISM} \citep{kirk2024prism} is especially relevant for a statistical treatment of RLHF. The dataset was designed to study how preferences over language model outputs vary across users and cultural settings. It contains large-scale human--AI interactions from 1,500 participants across 75 countries, covering responses from 21 different LLMs. In addition to preference or rating data, \texttt{PRISM} records user-level metadata such as demographic and sociocultural attributes. This makes it particularly useful for studying heterogeneous feedback, personalized reward modeling, and subgroup-specific alignment. In contrast to datasets that treat user variation as noise, \texttt{PRISM} makes such variation an explicit object of study.

Other datasets are also widely used in RLHF research. For example, \texttt{hh-rlhf} \citep{bai2022training} is an early and influential human-labeled pairwise preference dataset for helpful and harmless dialogue, while \texttt{UltraFeedback} \citep{cui2023ultrafeedback} uses large-scale AI-generated feedback to provide a broader and cheaper source of preference supervision. 

\subsection{RLHF Platforms}

To support reproducible and scalable RLHF training, several open-source frameworks have been developed. These platforms provide implementations of the main alignment stages, including supervised fine-tuning, reward modeling, and preference optimization.

Among them, \texttt{TRL} \citep{vonwerra2020trl} is one of the most widely used libraries for post-training language models. It supports methods such as PPO, DPO, and GRPO within the Hugging Face ecosystem. Other commonly used frameworks include \texttt{trlX} \citep{havrilla-etal-2023-trlx}, which provides infrastructure for RLHF training, and \texttt{DeepSpeed-Chat} \citep{yao2023deepspeed}, which implements an end-to-end large-scale RLHF pipeline.

To make these ideas concrete, we provide a small demonstration at \url{https://github.com/Pangpang-Liu/RLHF_demo}, built on \texttt{TRL}, that illustrates core RLHF training steps under both PPO and DPO.

\subsection{Evaluations}

Evaluation remains one of the hardest parts of RLHF because alignment is multi-dimensional: models may differ in correctness, helpfulness, safety, reasoning, factuality, tone, and cultural appropriateness. Human evaluation remains the gold standard, but it is expensive and difficult to scale. As a result, many modern benchmarks rely partly or fully on LLM-as-a-judge protocols \citep{gu2024survey}.

For this survey, the most relevant evaluation paradigm is arena-style pairwise comparison. In this setting, two models respond to the same prompt, and either human judges or strong LLM judges choose the preferred output. If model $m$ produces response $\boldsymbol y^{(m)}$ and model $m'$ produces response $\boldsymbol y^{(m')}$ for prompt $\boldsymbol x$, then one may model the comparison outcome as
$
\mathbb P\big(m \succ m'\mid \boldsymbol x\big)
=
\sigma\!\left(\Delta_{m,m'}(\boldsymbol x)\right),
$
where $\Delta_{m,m'}(\boldsymbol x)$ is a prompt-specific utility gap. Classical BTL models arise as a special case when this gap is prompt independent.

This view makes clear that LLM evaluation is closely connected to the same pairwise-comparison framework used in RLHF training. It also exposes several statistical challenges: comparison designs are highly unbalanced, model quality is prompt dependent, and judge behavior may itself be biased or heterogeneous. As a result, leaderboard construction naturally raises questions of contextual modeling, uncertainty quantification, calibration, and multiple comparisons. Existing systems such as \texttt{Prometheus-Eval}, Preference Proxy Evaluation, \texttt{AlpacaEval}, and \texttt{MT-Bench-101} illustrate the diversity of evaluation protocols now in use \citep{kim2024prometheus,frick2025how,alpaca_eval,bai2024mt}. For statisticians, the key message is that modern LLM evaluation is not separate from RLHF, but rather another important setting for pairwise preference modeling and inference.

\section{FUTURE DIRECTIONS AND OPEN PROBLEMS}\label{sec:challenge}

We conclude by highlighting several emerging directions that may warrant further attention. From a statistical perspective, progress in RLHF may depend not only on improved estimation and optimization but also on deeper consideration of issues such as privacy, fairness, safety, and governance.

\begin{itemize}[leftmargin=1.5em]

\item \textbf{Privacy and Data Protection.} 
RLHF often relies on sensitive preference data, user interactions, and annotator metadata. This raises privacy questions at several levels: protection of raw human feedback, privacy leakage through aligned models, and the tradeoff between privacy guarantees and alignment quality. Recent work has begun to study differentially private and robust versions of offline RLHF and DPO, showing that privacy constraints can materially affect estimation error and alignment performance \citep{cho2026privacy}. At a broader systems level, recent privacy guidance for LLMs also emphasizes data protection by design and risk management for deployment \citep{zhou2025unified,zhou2025squarepo}. These developments suggest an important open problem for statistics: how to quantify privacy-utility tradeoffs in preference learning and policy optimization, especially under adaptive querying and heterogeneous users. 

\item \textbf{Fairness, Pluralism, and Population Alignment.} 
Another emerging challenge is fairness. Standard RLHF often aggregates preferences into a single reward signal, but this can systematically favor majority groups or dominant value systems. Recent work on reward fairness in RLHF proposes fairness-aware regularization and reward adjustments to promote more equitable alignment outcomes, while broader studies of preference alignment argue that models should go beyond a single notion of harmlessness or helpfulness and better reflect pluralistic values \citep{ouyang2025rewardfairness}. Related empirical work also shows that preference alignment does not automatically improve all trustworthiness dimensions and may even worsen some of them, including stereotypical bias and privacy behavior \citep{li2025rlhfmoretrust}. For statisticians, this raises fundamental questions about estimands: should RLHF target average preferences, subgroup-specific preferences, Pareto-efficient compromises, or explicitly fairness-constrained objectives?

\item \textbf{Safety Guarantees and High-Confidence Alignment.} 
Most current RLHF methods optimize average preference-based objectives, but safety-critical applications may require stronger guarantees than average-case improvement. Recent work on high-confidence safe RLHF seeks to provide probabilistic guarantees that harmful responses remain below a target threshold while preserving helpfulness \citep{chittepu2025hcrlhf}. This points to a broader research agenda at the intersection of RLHF and statistical decision theory: how should one construct confidence-calibrated alignment procedures, certify safe deployment under uncertainty, and balance utility against risk in high-stakes settings? Such questions are likely to become increasingly important as aligned LLMs are deployed in domains such as healthcare, education, and decision support.

\item \textbf{Evaluation, Governance, and Auditing.} 
A final challenge is how aligned models should be evaluated and audited after training. Earlier sections discussed uncertainty-aware evaluation and arena-style pairwise comparison, but future work may also need post-deployment monitoring, subgroup auditing, and accountability benchmarks. Recent work moves in this direction: \citet{ojewale2026audit} propose audit trails for LLMs, while \citet{dong2026evaluating} show that combining labeled outcomes with pairwise signals can improve statistical efficiency and ranking precision. Since aligned models may behave differently across domains, populations, and trustworthiness dimensions, evaluation is not just a one-time benchmark but an ongoing monitoring problem. Developing methods for continual auditing, subgroup-aware monitoring, efficient comparative evaluation, and transparent reporting remains an important open direction.

\end{itemize}

{\baselineskip=22pt
\bibliographystyle{asa}
\bibliography{reference}
}

\end{document}